\documentclass[10pt, a4paper]{article}

\usepackage{lrec-coling2024} 

\usepackage[T1]{fontenc}
\usepackage[utf8]{inputenc}

\usepackage{times}
\usepackage{latexsym}

\usepackage{microtype}

\usepackage[group-separator={,}]{siunitx}

\usepackage{graphicx}
\usepackage{amsmath}
\usepackage{amssymb} 

\usepackage{booktabs}

\usepackage{caption}
\usepackage{subcaption}

\usepackage{enumitem} 
\usepackage{tabularx}
\usepackage{multirow}
\usepackage{listings}
\usepackage{multicol}

\setlist{nolistsep} 
\usepackage{amsfonts,dcolumn}

\usepackage{fancyhdr}
\usepackage{pifont}

\name{Ulme Wennberg, Gustav Eje Henter} 

\address{Division of Speech, Music and Hearing, KTH Royal Institute of Technology \\
         Lindstedtsv\"agen 24, SE-100 44 Stockholm, Sweden\\
         ulme@kth.se, ghe@kth.se\\}

\title{Exploring Internal Numeracy in Language Models:\\{}A Case Study on ALBERT}

\abstract{
It has been found that Transformer-based language models have the ability to perform basic quantitative reasoning. In this paper, we propose a method for studying how these models internally represent numerical data, and use our proposal to analyze the ALBERT family of language models. Specifically, we extract the learned embeddings these models use to represent tokens that correspond to numbers and ordinals, and subject these embeddings to Principal Component Analysis (PCA). PCA results reveal that ALBERT models of different sizes, trained and initialized separately, consistently learn to use the axes of greatest variation to represent the approximate ordering of various numerical concepts. Numerals and their textual counterparts are represented in separate clusters, but increase along the same direction in 2D space. Our findings illustrate that language models, trained purely to model text, can intuit basic mathematical concepts, opening avenues for NLP applications that intersect with quantitative reasoning.
\\ \newline  }

\begin{document}

\maketitleabstract

\begin{figure*}[t!]
  \centering
  \begin{subfigure}[b]{0.49\textwidth} 
    \includegraphics[width=\linewidth]{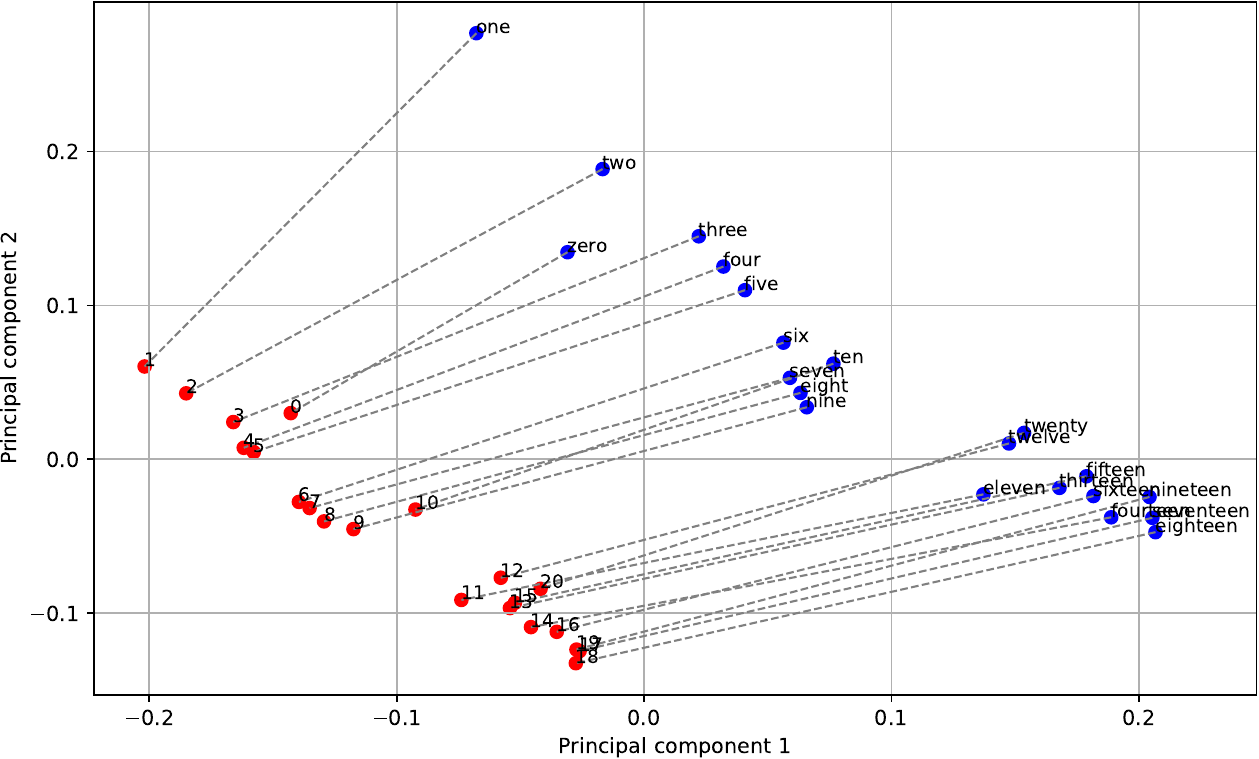}
    \caption{albert-base-v2}
    \label{fig:numeric_and_word_pairs_basev2}
  \end{subfigure}
  \hfill 
  \begin{subfigure}[b]{0.49\textwidth} 
    \includegraphics[width=\linewidth]{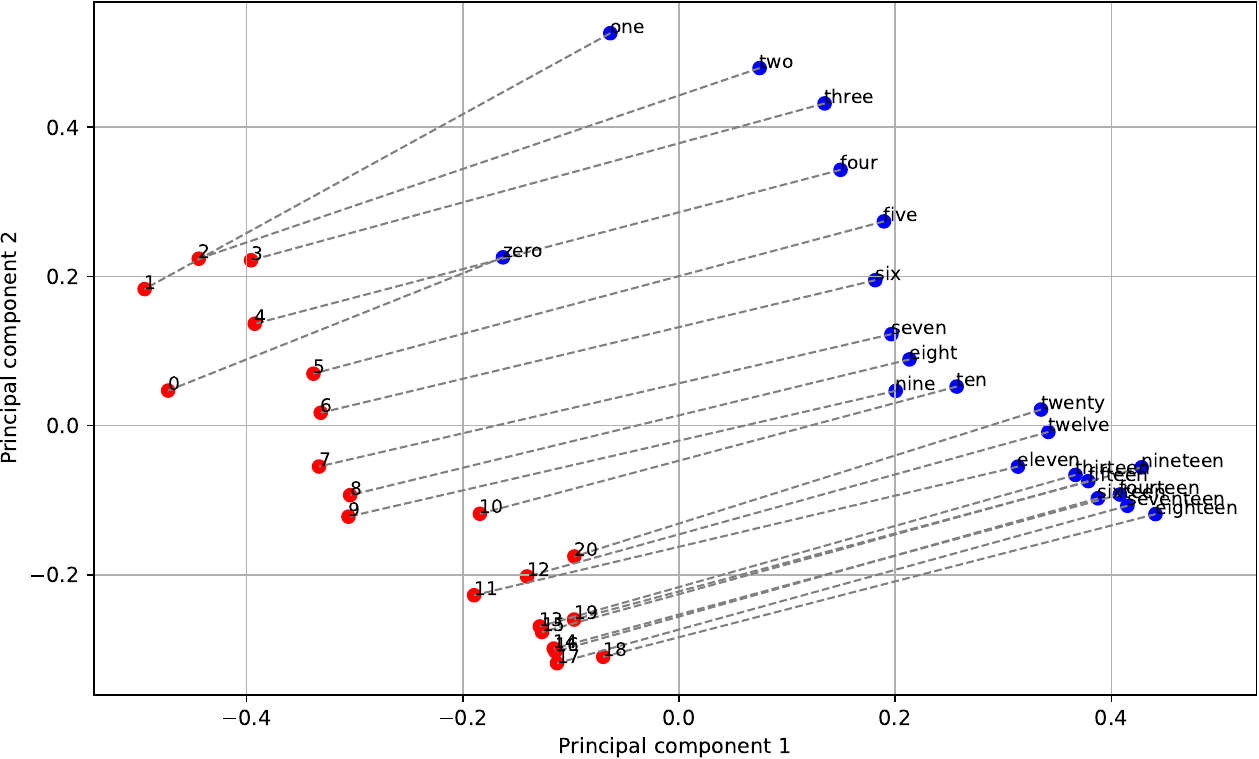}
    \caption{albert-xxlarge-v2}
    \label{fig:numeric_and_word_pairs_xxlargev2}
  \end{subfigure}
  \vspace{-1mm}
  \caption{Visualization of the two first principal components of word embeddings for numbers zero through twenty and their textual counterparts in two ALBERT models.}
  \vspace{-1mm}
  \label{fig:numeric_and_word_pairs}
\end{figure*}%

\section{Introduction}
\label{sec:intro}
The Transformer architecture introduced by \citet{vaswani2017attention} has led to major advances in computational linguistics.
Transformer-based models of language like BERT \citep{devlin2019bert}, RoBERTa \citep{liu2019roberta}, ALBERT \citep{lan2020albert}, and ELECTRA \citep{clark2020electric} have excelled in a range of tasks, from machine translation \citep{lample2018phrasebased, gu2018metalearning} to question answering \citep{yamada2020luke} and beyond.
Studies have also evaluated these models' numerical reasoning,
For example, \citet{saxton2019analysing} found a Transformer-based language model to perform at an E-grade level on a British math exam for 16-year-olds.
\citet{kalyan2021coffee} subsequently developed benchmarks for mathematical commonsense reasoning to track the progress of models in this respect, highlighting the growing interest in this research area.

Despite demonstrable performance on numerical tasks, the origin of these abilities in text-trained models, and the root of their numeracy and quantitative understanding, remain obscure.
In this paper, we aim to illuminate this aspect by analyzing the learned embeddings these models use to represent lexical tokens internally.
We study how the models have learned to embed numerals and their written representations, unearthing evidence that various embedding vectors capture the essence of numerical concepts.
Instead of studying whether the embeddings of different numbers are distributed close together (like one might cluster the embeddings of synonyms; cf.\ \citet{mikolov2013distributed}), we thus consider how their representations \emph{differ}, and how the axes of greatest variation among these concepts relate to the intrinsic ordering and numeric value of the different tokens.

Our paper makes two contributions:
\begin{enumerate}
\item We propose a novel way to study internal numerical cognition in language models.
\item We use our proposed method to investigate how ALBERT encodes numerical and ordinal information, and how this varies across different versions of ALBERT of various sizes, independently trained.
\end{enumerate}
Using our proposed method, we find:
\begin{itemize}
\item
Trained ALBERT models consistently use primary principal component axes to denote ordering and spacing of numbers, ordinals, and magnitude orders.
\item The representations are closer together for higher values, suggesting a logarithmic representation of numbers.
\item Numerals and their textual counterparts are represented in separate clusters, but increase along the same direction in 2D PCA space.
\end{itemize}

\vspace{-2mm}
\section{Background}
\label{sec:background}
In this section, we give a background on numeracy in language models and highlight previous works investigating their internal token representations.

Numeracy is critical for complex reasoning in NLP. Investigations into models' numerical reasoning abilities \citep{wallace2019nlp, jin2021numgpt, thawani2021numeracy, duan2021learning, sakamoto2021predicting} have shown promising enhancements in model numeracy. Further studies \citep{kim2021seen, lin2020numeralsense, shah2023numeric} have explored models' abilities to extrapolate and their numerical commonsense knowledge. In conjunction with these explorations, recent methodologies \citep{sundararaman2020numbers, saeed2022type, jiang2020learning, liang2022mwpbert} introduce a variety of approaches to improve numerical representation and processing, showing ongoing efforts to refine numeracy in language models.

Internally, Transformers use self-attention to capture dependencies between all pairs of input vectors.
The models use a variety of mechanisms to represent positional information for sequential input data.
One common approach, used already in \citet{vaswani2017attention}, is to encode each token $w_i$ as a vector that is the sum of a content-embedding vector (that depends only on the token $w_i$) and a position-encoding vector (that depends only on $i$, the position in the input sequence).
The position-dependence mechanisms either explicitly represent the inherent ordering of input symbols, or, when they do not, have been shown to learn positional representations that capture both this ordering and the translation equivariance of text sequences \citep{wennberg2021tisa}.

Other research (e.g., \citealp{mikolov2013distributed, vylomova2016take, durrani-etal-2022-transformation}) has sought to shed light on information processing in neural language models by analyzing their learned embeddings of different words, concepts, and lexical tokens.
A consistent finding is that synonyms cluster together in latent space, meaning that linguistic similarity is reflected internally in the learned model.
In this work, we apply a similar analysis, but to concepts that are numerical rather than linguistic.
The key difference between our present study and prior work on language-model numeracy is that we look directly at the internal embeddings that Transformer-based language models have learned for numerical concepts, and investigate to what extent \emph{differences} between these embeddings are reflective of differences in numerical value between the mathematical concepts they represent.

\vspace{-2mm}
\section{Experiments}
\label{sec:experiments}

We now describe the method and results of our study of the word embeddings inside eight different Transformer-based language models, namely the ALBERT family \citep{lan2020albert}.
We choose to study ALBERT because it is available in four different model sizes (starting at ``base'' and going up to ``xxlarge''), each with checkpoints at two different points during training (``v1'' vs.\ ''v2'', with v2 having been trained for longer), allowing for a comparison of embeddings in different models and their evolution.
In our analysis, we specifically examine numerical ranges from zero to twenty and one to one hundred, not only to cover a broad spectrum of basic and multi-digit numerals but also because these numbers are consistently tokenized as single tokens by the ALBERT models \citep{lan2020albert}. This choice aligns with our objective to study unambiguous, uncontextualized numerical representations within the model.
Many submissions to the GLUE \citep{wang2019glue} and SuperGLUE \citep{wang2020superglue} leaderboards are descendants of the ALBERT architecture.

\begin{figure*}[t!]
  \centering
  \begin{subfigure}[b]{0.49\textwidth} 
    \includegraphics[width=\linewidth]{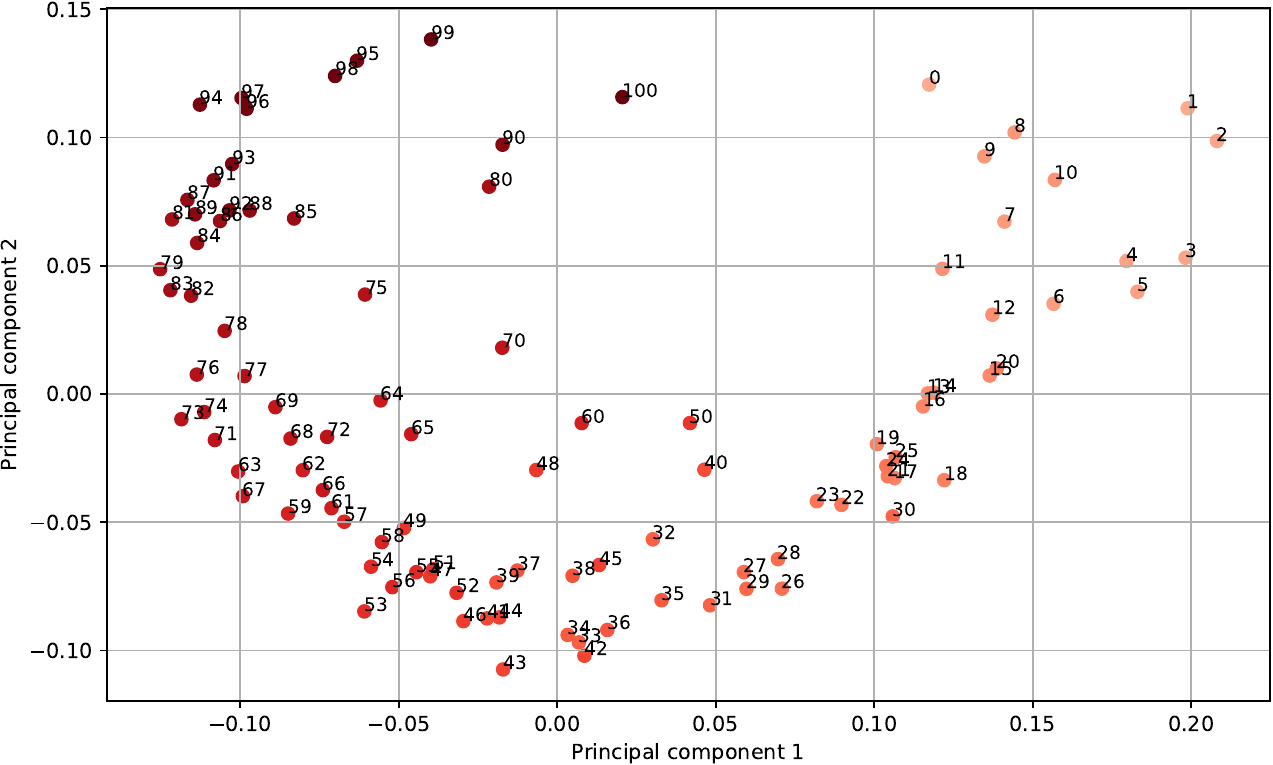}
    \caption{albert-base-v2}
    \label{fig:fig_all_numbers_to_100_albert-base-v2}
  \end{subfigure}
  \hfill 
  \begin{subfigure}[b]{0.49\textwidth} 
    \includegraphics[width=\linewidth]{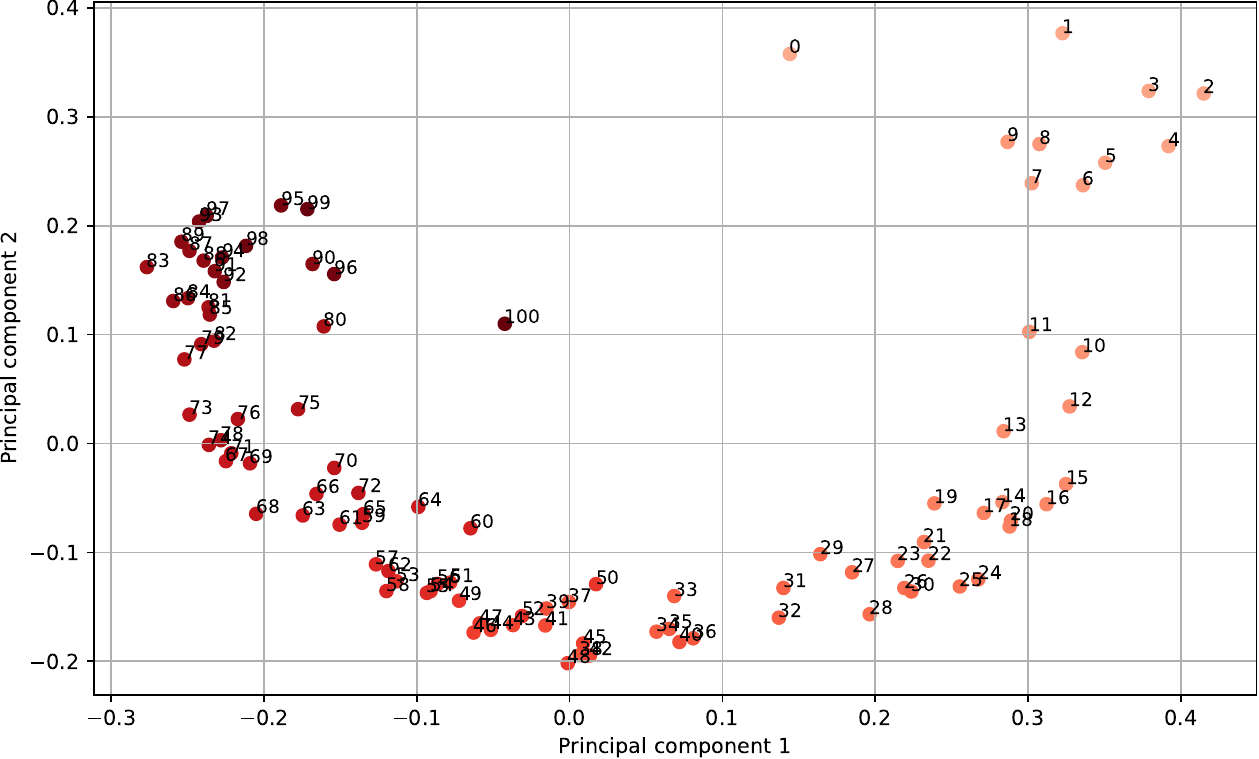}
    \caption{albert-xxlarge-v2}
    \label{fig:fig_all_numbers_to_100_albert-xxlarge-v2}
  \end{subfigure}
  \vspace{-1mm}
  \caption{The first and second PCA components for all numbers 1 to 100 in two different ALBERT models.}
  \vspace{-1mm}
  \label{fig:all_numbers_to_100}
\end{figure*}%

\subsection{Analysis Methodology}
All the analyses in this paper follow the same underlying recipe:

First, we extract uncontextualized embeddings for selected tokens (single-token words only). These word embeddings are prior to position embedding addition or self-attention layer processing.

We then conduct PCA on these embeddings to identify principal variation axes.
This is a linear dimensionality reduction technique, meaning that linear structures like a number line are preserved.

Lastly, we plot embeddings along principal component axes to assess if they capture mathematical concept ordering and if distances reflect mathematical relationships, like proximity of similar numbers.

\subsection{Numerical vs.\ Lexical Embedding}
We first compare the learned embeddings of the numbers zero through twenty, juxtaposed with their word representations (e.g., ``7'' versus ``seven''), across different ALBERT models.
The results of applying our PCA-based analysis to the resulting 42 different representations is visualized in Figure \ref{fig:numeric_and_word_pairs} for the smallest and largest ALBERT models (the other models yield very similar-looking plots).

A number of observations are immediately apparent in the figure:
\begin{enumerate}
\item Numbers and words representing occupy two distinct, elongated clusters.
\item Within each cluster, there is a direction along which numerical values generally increase. In other words, the values are mostly in order, and we (approximately) recover a number line in the PCA space for each cluster (numbers vs.\ number words).
\item The direction along which the values increase is the same for both numbers and number words. It would thus easily be possible, particularly for the bigger model, to project the embeddings onto a single axis in PCA space that approximates the number line.
\item When values exceed ten, numbers begin to bunch up more.
\end{enumerate}
The fact that the two different kinds of embeddings can be projected onto something like the number line strongly suggests a learned ability to link numerical symbols to their word forms, and to their approximate value and ordering.

We can also make some minor observations about individual numbers, such as the positions of the numbers and words for zero being idiosyncratic, and (more curiously) that numbers and words for twenty also consistently are out of place.

\subsection{Numbers 1 Through 100}
Next, we performed the same analysis on integers 0 to 100 (excluding word forms) and charted their 2D PCA distribution for the same two ALBERT models. The findings, displayed in Figure \ref{fig:all_numbers_to_100}, mirror those from other models.
%
We observe that:
\begin{enumerate}
\item As numbers increase, they approximately trace out a horseshoe shape in 2D space.
\item Larger numbers gradually compress closer together, especially for the larger model.
\item Rounded numbers (i.e., those ending in zero) lie closer to middle of the space. This is more visible for the smaller model, but true for both. 25, 75, and numbers with many powers of two are also closer to the middle. 100, with two zeroes, sticks out particularly much.
\end{enumerate}
The most important conclusion is that the ability to use embeddings to order numbers by size persists into larger numbers, though the spacing gets more compressed as the numerical values increase.

\subsection{Representing Orders of Magnitude}
\begin{figure}[t!]
\centering
\includegraphics[width=\linewidth]{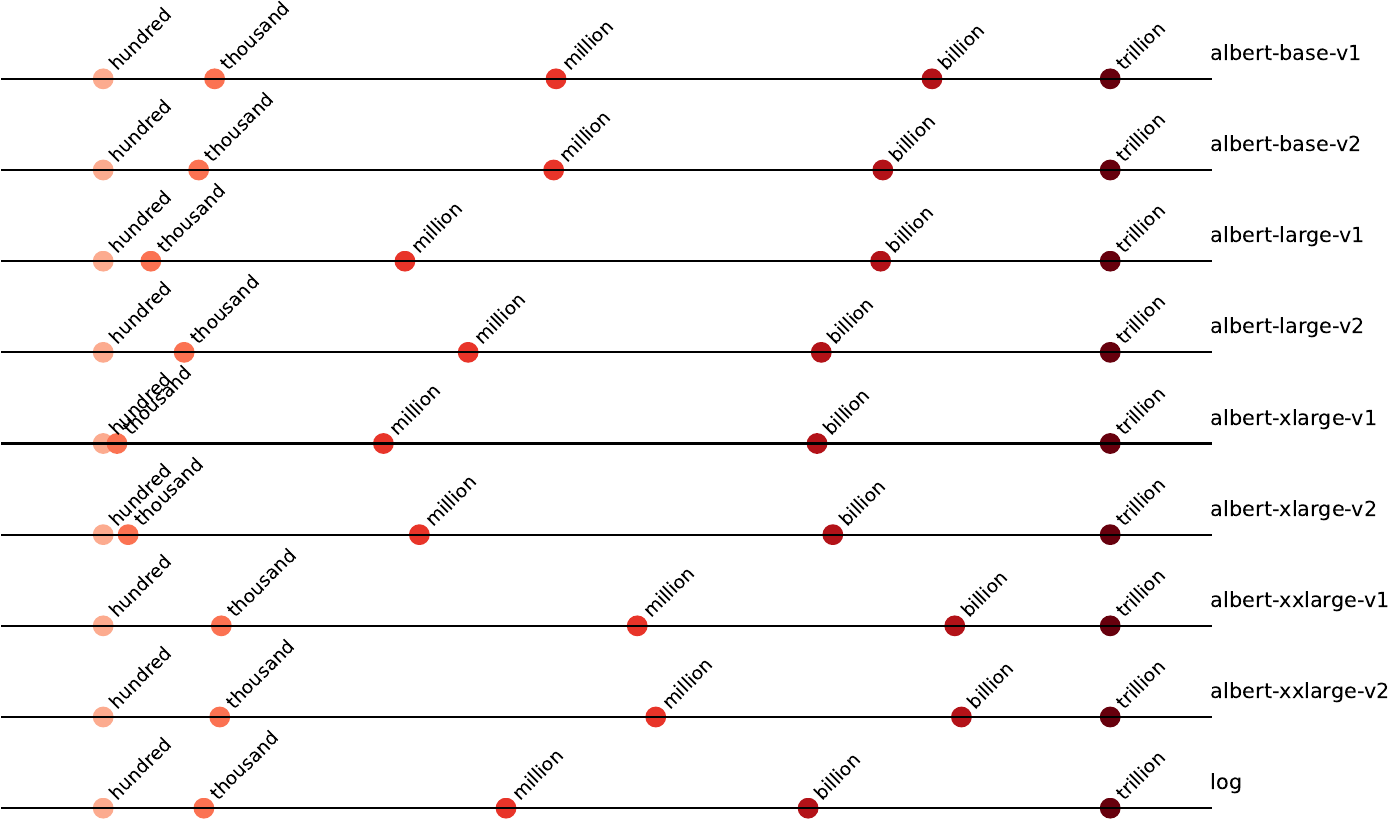}
  \vspace{-1mm}
\caption{Orders-of-magnitude word embeddings visualized along the first PCA axis across eight ALBERT configurations. Axes have been affinely transformed so that the first and last embeddings line up vertically. The last row shows the concepts arranged on a logarithmic axis for comparison.}
  \vspace{-1mm}
\label{fig:large_scale_numbers}
\end{figure}

Having looked at numbers up to 100, we also studied the embeddings of words for different orders of magnitude.
Specifically, we performed PCA on the embedding representations of the words ``hundred,'' ``thousand,'' ``million,'' ``billion,'' and ``trillion.''
Figure \ref{fig:large_scale_numbers} shows these words' positions on the first principal axis across eight ALBERT models.
%
We see that:
\begin{enumerate}
\item The words always respect the expected ordering based on their numerical value.
\item The separation between ``hundred'' and ``thousand'' is consistently the shortest, typically by some margin. This evokes comparisons to the logarithmic axis at the bottom of the figure.
\end{enumerate}
There is a close call between ``hundred'' and ``thousand'' for the xlarge model, but the separation increases with longer training (model v1 vs.\ v2).

\subsection{Words for Ordinals}
\begin{figure}[t!]
  \centering
  \includegraphics[width=\linewidth]{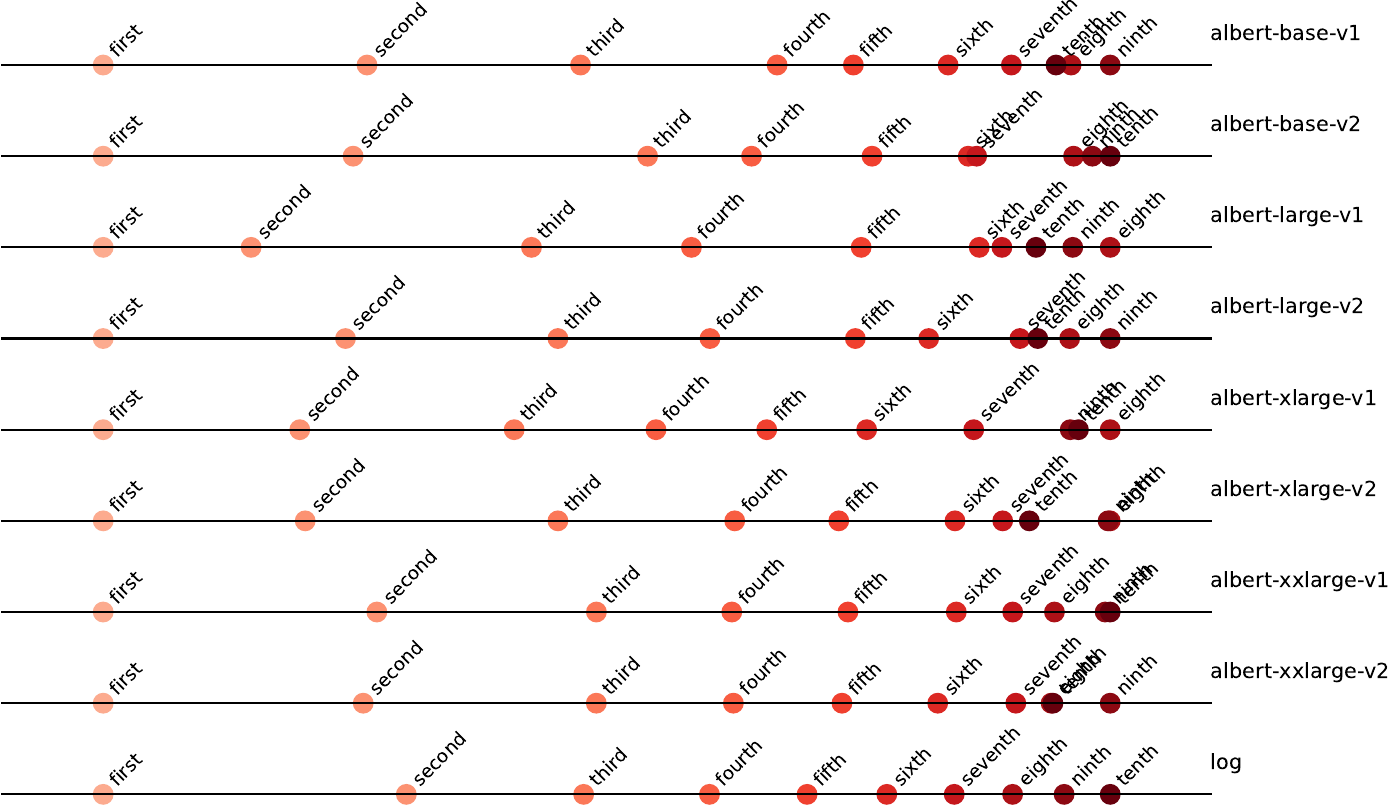}
  \vspace{-1mm}
  \caption{Visualization of ordinal term embeddings along the first PCA axis across eight ALBERT configurations. The axes have been affinely transformed so that the first and last embeddings line up vertically. The last row shows the concepts arranged on a logarithmic axis for comparison.}
  \vspace{-1mm}
  \label{fig:ordinal_numbers}
\end{figure}

As our last experiment, we visualize the representation of words for ordinals rather than numerals.
Specifically, we apply the same PCA-based method to the embeddings of the terms ``first'' through ``tenth'' and visualize the first principal axis for all eight ALBERT models, like in the previous section.
The results are shown in Figure \ref{fig:ordinal_numbers}, from which we make the following observations:
\begin{enumerate}
\item Ordinals consistently appear in the correct order along the principal axis of variation up until and including ``seventh''.
\item The distance between ordinals gradually decreases as the numbers increase, with the last three ordinals generally being close together and often out of order.
\end{enumerate}
Embeddings do not become obviously better with longer training, especially as they already mostly appear in the correct order for the v1 models.

\vspace{-2mm}
\section{Discussion}
Reviewing all four analyses, it's evident that ALBERT models' internal representation of various numbers and numerical concepts in the embedding layer directly reflects their numerical value.
The representations are closer to a logarithmic than a linear scale.
These trends are very consistent across models of different sizes and trained for different amounts of time.

While the fact that Transformer-based language models can support simple mathematical reasoning has indicated some level of numeracy within the model, we can now open the black box and see that numerical knowledge evident within the basic vector representations inside the model.
It is not at all obvious that reasonable representations of numerical concepts would arise in these models, given that they are pre-trained exclusively on text to optimize standard language-modeling objectives, with no direct mathematical training.

The observation that larger numbers cluster closer, hinting at logarithmic scaling, and the unique behavior of round numbers in Figure \ref{fig:all_numbers_to_100}, may stem from their occurrence frequency in data, aligning with Benford's law \citep{benford1938law}. This law suggests that smaller leading digits are more common in real-world numerical data, yielding a near uniform distribution of digits on a logarithmic scale.

A notable limitation of our study is its focus on single-token numbers, which excludes decimals and larger numerical values from our analysis.

\vspace{-2mm}
\section{Conclusion}
We have introduced a novel approach to analyzing the quality of numerical representations in language models.
This offers insights into model numeracy, which matters for developing improved numerical-understanding capabilities for Transformer-based language models.

We use our method to investigate how ALBERT, an important Transformer-based language model architecture, represents different numerical and ordinal inputs.
Our results demonstrate a clear concept of numerical ordering within the vector representations inside the model.
Representations of larger numbers fall closer together, suggestive of models using logarithmic axis representations internally.
The findings are very robust, in that they appear essentially unchanged across eight different models that differ in size and training duration.

Going beyond numerical order, future work should seek to quantify to what extent learned internal structures reflect interval and ratio scales, as well as to what extent factors like model architecture and term frequency in the corpus contribute to (or otherwise influence) these structures.
Another goal is to extend the analysis to multi-token numbers and mathematical operators, and connect with emerging understanding of how models then perform stepwise mathematical processing in latent space \citep{lee2019mathematical,valentino2023multioperational}.

\section{Acknowledgments}
We thank the anonymous reviewers for constructive feedback.
This research was partially supported by the Wallenberg AI, Autonomous Systems and Software Program (WASP) funded by the Knut and Alice Wallenberg Foundation.

\section{Bibliographical References}
\bibliographystyle{lrec-coling2024-natbib}
\bibliography{lrec-coling2024}

\end{document}